\documentclass[runningheads]{llncs}
\usepackage[T1]{fontenc}
\usepackage{graphicx}
\usepackage{booktabs}
\usepackage{amsmath}
\usepackage{amssymb}
\usepackage{siunitx}
\usepackage{xcolor}
\usepackage{tikz}
\usetikzlibrary{arrows.meta, positioning, shapes.geometric, fit, backgrounds, calc}
\usepackage[hidelinks]{hyperref}

\usepackage{placeins}   
\begin{document}
\title{SADL: A Size-Adaptive Decision Layer for Chronic Stroke Lesion Segmentation in ISLES~2026}
\titlerunning{SADL for ISLES 2026}
\author{Marek Wodzinski\inst{1,2}\orcidID{0000-0002-8076-6246} \and
Gniewosz Drwiega\inst{1}\orcidID{0000-0002-1968-2238} \and
Wojciech Szymanski\inst{1,2}\orcidID{0009-0001-2043-9104}}
\authorrunning{M. Wodzinski et al.}
\institute{Sano -- Centre for Computational Personalised Medicine International
Research Foundation, Krakow, Poland \and
AGH University of Krakow, Krakow, Poland\\
\email{wodzinski@agh.edu.pl}}
\maketitle
\begin{abstract}
Chronic ischemic stroke lesions are difficult to segment on T1-weighted
MRI. Their appearance overlaps with healthy tissue, and their volumes span
four orders of magnitude, so voxel-level and lesion-level metrics often
disagree. We build an eleven-member nnU-Net ensemble that covers five design
axes: a baseline dice + top-K loss, heavy augmentation, per-lesion-volume
loss re-weighting, stochastic weight averaging, and a
sampling-and-architecture block that pairs cohort-balanced sampling with
a residual-encoder backbone. Weighted soft-map fusion of the members feeds a tuned decision layer,
\emph{SADL}, that estimates lesion volume, applies an adaptive
per-lesion-size threshold, prunes components by voxel count and peak
probability, and adds a never-empty rescue. All fusion weights and
per-bucket parameters are fitted on out-of-fold (OOF) predictions, so
tuning never touches the held-out test set. On a site-disjoint
91-case holdout, SADL reaches Dice~0.7322, lesion-F1~0.7461 and
PR-AUC~0.8430 when scored offline with the challenge's released
evaluation code. A paired non-parametric bootstrap against a reference
single ensemble member confirms the lesion-F1 gain at
$p{=}0.024$, and the gain is largest in the small-lesion regime, where
the rescue step recovers cases that a fixed threshold would silently
discard.
\end{abstract}
%
\section{Introduction}
\label{sec:intro}

Chronic stroke lesion segmentation on T1-weighted MRI feeds directly into
outcome prediction, neuro-restorative studies, and post-stroke rehabilitation
planning. Manual delineation is slow and rater-variable, and the task itself
is genuinely hard: chronic lesions look very close to normal tissue, their
volumes span from a fraction of a millilitre to several hundred, and public
cohorts are pooled across scanners with different acquisition protocols. In
that regime, voxel-level metrics such as Dice can overstate performance on
cases whose small lesions dominate lesion-wise recognition~\cite{kofler2023panoptica}.

Since the release of ATLAS~v2, most published models for this task have been
variations around nnU-Net~\cite{isensee2021nnunet,isensee2024nnunet_revisited}
and its residual-encoder update. Once the backbone plateaus around Dice~0.72
on ATLAS-derived holdouts, further gains typically come from three levers:
training-side variants (loss re-weighting, heavier augmentation, weight-space
averaging, cohort balancing), architectural diversity (residual and
transformer-hybrid encoders), and post-hoc calibration of the decision layer
that turns soft ensembles into binary masks.

This paper works on the third lever. We assemble an eleven-member nnU-Net
ensemble that covers five design axes and, on top of a weighted soft-map
fusion of its outputs, add a decision layer we call \emph{SADL}. The
layer estimates lesion volume, applies an adaptive per-lesion-volume
threshold, prunes connected components, and adds a never-empty rescue. Every fusion weight and per-bucket parameter is fitted
on out-of-fold predictions inside a 5-fold cross-validation, so the 91-case
site-disjoint holdout is only used for reporting. Under the ISLES~2026
released evaluation code~\cite{kofler2023panoptica}, the tuned layer
beats the Dice-strongest single ensemble member (BucketWeighted,
$v_{\text{ref}}=3$~mL) by $+0.6$~pp Dice and $+5.4$~pp lesion-F1
overall, and by $+7.4$~pp lesion-F1 on the $<0.5$~mL bucket; a paired
bootstrap against a separate reference member ($v_{\text{ref}}=5$~mL)
gives a lesion-F1 lift of $+2.9$~pp at $p{=}0.024$, and the improvement
concentrates in the small-lesion regime where the rescue step earns its
keep.

\section{Methods}
\label{sec:methods}

\subsection{Dataset and splits}
Training data comes from ATLAS~v2.0~\cite{liew2022atlas} together with
the Stroke Outcome Optimization Project (SOOP,
169~sessions)~\cite{absher2024soop}, aggregated and released through
ISLES~2026~\cite{isles2026dataset}: 1{,}453 T1-weighted sessions across
55 sites, each with a manually delineated binary lesion mask. The pool
is built from three main cohorts (Training: 653 sessions; ATLAS3, which
subsumes SOOP: 496; Testing: 299) plus a residual of five sessions with
missing or auxiliary cohort labels, absorbed into the main cohorts using
their modal label. We reserve three sites (R018, R027, R047) as a
site-disjoint 91-case holdout that is large enough for a per-lesion-size
analysis while still fitting the CPU budget of our decision-layer tuner,
and we run 5-fold cross-validation on the remaining 1{,}362 sessions. All models are
trained in nnU-Net's plans space at $1\,\mathrm{mm}$ isotropic; predictions
are resampled back to native T1w geometry before scoring.

\subsection{Backbone}
Every member of our pool is a subclass of nnU-Net v2.7.0~\cite{isensee2021nnunet}.
The upstream codebase is kept unmodified and all our extensions live in a
thin wrapper package, so training and inference follow nnU-Net's established
recipe (deep supervision, Gaussian sliding-window inference, mirror-based
test-time augmentation) unless a specific member deliberately overrides it.
Two encoder families appear in the pool: the default PlainConvUNet, and the
residual-encoder variant from the ``nnU-Net Revisited''
update~\cite{isensee2024nnunet_revisited}.

\subsection{Ensemble pool}
The eleven members span five design axes chosen to inject complementary
error modes: a \emph{baseline} (dice-plus-top-K cross-entropy); an
\emph{augmentation} member (DA5 heavy schedule, 700 epochs); a
\emph{loss re-weighting} axis (\emph{BucketWeighted}) with per-sample
weight
$w_i = \operatorname{clip}(v_{\text{ref}} / \max(v_i, \varepsilon), 0.5, 4.0)$
at three reference volumes $v_{\text{ref}} \in \{2, 3, 5\}$~mL;
\emph{weight-space averaging} (SWA~\cite{izmailov2018swa} from
epoch~350, combined with the loss re-weighting); and a
\emph{sampling-and-architecture} block gathering cohort-balanced
sampling, curriculum-by-lesion-size, a cohort-conditioned
mixture-of-experts head, synthetic-lesion diffusion augmentation, and
the ResEnc backbone. All eleven members are trained
independently on the standard nnU-Net 5-fold split of the training pool,
and every fold's raw soft-max is written to disk so that the decision
layer can be tuned offline.

\subsection{SADL decision layer}
\label{sec:pillar1}
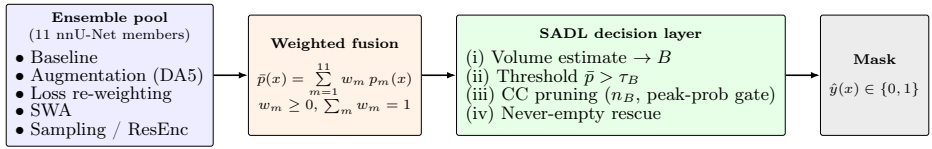
\begin{figure}[t]
\centering
\resizebox{\textwidth}{!}{%
\begin{tikzpicture}[
    node distance=6mm,
    every node/.style={font=\scriptsize},
    box/.style={draw, rounded corners=1.2pt, align=center, inner sep=3pt},
    poolbox/.style={box, fill=blue!7,   minimum width=28mm, minimum height=21mm},
    fusebox/.style={box, fill=orange!10, minimum width=22mm, minimum height=21mm},
    sadlbox/.style={box, fill=green!10, minimum width=32mm, minimum height=21mm},
    maskbox/.style={box, fill=gray!15,  minimum width=18mm, minimum height=21mm},
    arr/.style={-{Latex[length=1.6mm]}, thick},
]

\node[poolbox] (pool)
    {\begin{tabular}{c}
        \textbf{Ensemble pool}\\
        {\scriptsize (11 nnU-Net members)}\\[3pt]
        \begin{tabular}{@{}l@{}}
        \small $\bullet$ Baseline\\
        \small $\bullet$ Augmentation (DA5)\\
        \small $\bullet$ Loss re-weighting\\
        \small $\bullet$ SWA\\
        \small $\bullet$ Sampling / ResEnc\\
        \end{tabular}
     \end{tabular}};

\node[fusebox, right=6mm of pool] (fuse)
    {\begin{tabular}{c}
        \textbf{Weighted fusion}\\[4pt]
        $\bar p(x) = \sum\limits_{m=1}^{11} w_m\, p_m(x)$\\[4pt]
        {\scriptsize $w_m \geq 0$, $\sum_m w_m = 1$}
     \end{tabular}};

\node[sadlbox, right=6mm of fuse] (sadl)
    {\begin{tabular}{c}
        \textbf{SADL decision layer}\\[3pt]
        \begin{tabular}{@{}l@{}}
        \small (i) Volume estimate $\to B$\\
        \small (ii) Threshold $\bar p > \tau_B$\\
        \small (iii) CC pruning ($n_B$, peak-prob gate)\\
        \small (iv) Never-empty rescue\\
        \end{tabular}
     \end{tabular}};

\node[maskbox, right=6mm of sadl] (out)
    {\begin{tabular}{c}
        \textbf{Mask}\\[6pt]
        $\hat y(x) \in \{0,1\}$
     \end{tabular}};

\draw[arr] (pool.east) -- (fuse.west);
\draw[arr] (fuse.east) -- (sadl.west);
\draw[arr] (sadl.east) -- (out.west);
\end{tikzpicture}%
}
\caption{Overview of the pipeline. Eleven nnU-Net members spanning five
design axes feed a weighted-mean soft fusion; the fused foreground field
enters the SADL decision layer, which estimates the lesion volume,
selects a per-bucket threshold, prunes low-confidence connected
components, and rescues cases that would otherwise collapse to an empty
prediction.}
\label{fig:pipeline}
\end{figure}

Given the eleven per-member foreground probability maps $p_m(x)$ for a case,
we first take a \emph{weighted fusion}
$\bar{p} = \sum_m w_m\, p_m$ with $w_m \geq 0$ and $\sum_m w_m = 1$, and
then feed $\bar{p}$ into the four-stage SADL decision layer
(Fig.~\ref{fig:pipeline}):
(i) a \emph{volume estimate} obtained by binarising $\bar{p} > 0.5$,
extracting 26-connected components, and reading the total foreground
volume $V^{\star}$;
(ii) an \emph{adaptive threshold} $\tau_B$ selected from a lookup table
indexed by the size bucket
$B(V^{\star}) \in \{<0.5, 0.5\text{--}5, 5\text{--}50, \geq 50\}\,\mathrm{mL}$;
(iii) \emph{connected-component pruning} that drops components below a
per-bucket voxel count $n_B$ or a per-CC peak-probability gate; and
(iv) a \emph{never-empty rescue} that reinstates the highest-probability
component whenever step (iii) leaves an empty mask, preventing
catastrophic zero predictions on very small lesions. The tunable
parameters ($w_m$, $\tau_B$, $n_B$, per-CC gates) are all fitted per
fold on the OOF validation set: weights by coordinate ascent on the
simplex against a Dice-plus-lesion-F1 composite, and per-bucket
parameters by a coarse grid followed by coordinate ascent. The tuner
runs in minutes on a modest CPU allocation, so we re-tune whenever a
new member joins the pool. The fitted thresholds sit
around $0.29$ on the two small-lesion buckets and $0.41$ on the mid and
large buckets: the small-lesion gate is deliberately more permissive to
compensate for the ensemble's under-confidence there.

\subsection{Evaluation}
Every configuration is scored offline with the evaluation code released
by the ISLES~2026 organisers alongside the challenge, built on top of
panoptica~\cite{kofler2023panoptica}. No challenge-server submission or
leaderboard number is involved. Dice is the panoptica global binary
Dice coefficient over the full foreground mask; lesion-F1 is
panoptica's recognition quality under naive greedy $1{:}1$ matching at
IoU~$\geq 0.25$; absolute lesion count and volume differences come from
the same evaluation; PR-AUC is computed on the fused foreground soft map.
We report the mean over the 91-case holdout and stratify by lesion size
(bucket populations range from 5 cases in $<0.5$~mL to 33 in $5$--$50$~mL).

\section{Results}
\label{sec:results}

Table~\ref{tab:main} compares the tuned SADL ensemble against a curated
subset of the eleven-member pool and against the uniform-mean-fusion baseline.
The table keeps one row per ensemble axis rather than listing every member,
since the axes motivate the pool and the individual differences within an
axis (for example, the three reference-volume variants of the
volume-weighted loss) are of secondary interest here.

SADL tops every metric except absolute volume difference. Against the
Dice-strongest single member (BucketWeighted with $v_{\text{ref}}=3$~mL)
the gain is $+0.62$~pp Dice and $+5.4$~pp lesion-F1; against the uniform-mean
ensemble the gain is $+1.03$~pp Dice, $+6.1$~pp lesion-F1, and $+0.14$~pp
PR-AUC. The absolute lesion count difference also drops from $0.98$ to
$0.86$, or about eleven fewer spurious components across the 91-case
holdout.

\begin{table}[h]
\caption{Overall ablation on the 91-case site-disjoint holdout, scored
offline with the ISLES~2026 released evaluation code. Higher is better
for Dice / lesion-F1 /
PR-AUC, lower is better for the two absolute-difference metrics.
Best per column in \textbf{bold}.}
\label{tab:main}
\centering
\small
\begin{tabular}{lccccc}
\toprule
Configuration & Dice~$\uparrow$ & LF1~$\uparrow$ & PR-AUC~$\uparrow$ &
$\Delta$Vol.~mL~$\downarrow$ & $\Delta$Count~$\downarrow$ \\
\midrule
Baseline (dice + top-K)                 & 0.7171 & 0.6746 & 0.8358 & 5.15 & 0.98 \\
ResEnc                                  & 0.7118 & 0.6731 & 0.8352 & 5.33 & 1.02 \\
DA5 (heavy augmentation)                & 0.7241 & 0.6743 & 0.8236 & 5.36 & 1.23 \\
BucketWeighted ($v_{\text{ref}}=3$~mL)  & 0.7260 & 0.6920 & 0.8351 & 5.35 & 0.96 \\
BucketWeighted $\times$ SWA             & 0.7212 & 0.7018 & 0.8304 & 5.32 & 0.98 \\
Uniform-fusion ensemble (top-11)        & 0.7219 & 0.6851 & 0.8416 & \textbf{5.13} & 0.98 \\
SADL tuned (ours)                       & \textbf{0.7322} & \textbf{0.7461} &
                                          \textbf{0.8430} & 5.47 & \textbf{0.86} \\
\bottomrule
\end{tabular}
\end{table}

The uniform-mean ensemble averages the eleven soft maps and thresholds
them at $0.5$ before a modest per-bucket clean-up, and it gains only
$+0.5$~pp Dice over the vanilla dice + top-K baseline. Almost the whole
SADL gain therefore comes from the decision layer, not from adding one
more ensemble member.

\begin{figure}[!ht]
\centering
\includegraphics[width=0.95\textwidth]{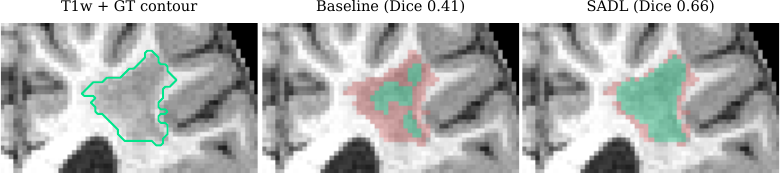}
\caption{Case at 5.4\,mL: SADL produces a tighter, more complete mask
than the baseline. GT contour and TP in green, FP in blue, FN in red.}
\label{fig:qual}
\end{figure}

Table~\ref{tab:bucket} slices the same 91 cases into four size buckets and
reports both Dice and lesion-F1. Two things stand out. SADL wins three of
the four buckets on Dice; on the $0.5$--$5$~mL bucket BucketWeighted
($v_{\text{ref}}=3$~mL) retains a small edge (0.596 vs 0.579). SADL's
margin over that BucketWeighted variant is small on $\geq 50$~mL cases
(0.877 vs 0.874) and largest on $<0.5$~mL cases ($+4.4$~pp).
Lesion-F1 is where the decision layer really earns its keep: at
$<0.5$~mL SADL scores $0.767$, about $7.4$~pp above the strongest
single-model config and $17.4$~pp above the uniform-mean ensemble. The
rescue step is why. Without it, a fixed $0.5$ threshold silently drops
the small-lesion cases where no voxel reaches the plurality vote, and
lesion-F1 collapses. SADL also tightens mid-size masks, as
Figure~\ref{fig:qual} illustrates on a $5.4$~mL case.

\begin{table}[h]
\caption{Per-lesion-size breakdown for Dice and lesion-F1. Bucket
populations: $<0.5$~mL: 5 cases, $0.5$--$5$~mL: 29, $5$--$50$~mL: 33,
$\geq 50$~mL: 24. Best per row and metric in \textbf{bold}.}
\label{tab:bucket}
\centering
\small
\setlength{\tabcolsep}{4.5pt}
\begin{tabular}{l cc cc cc cc}
\toprule
& \multicolumn{2}{c}{Baseline} & \multicolumn{2}{c}{Bwt-3} &
  \multicolumn{2}{c}{Uniform ens.} & \multicolumn{2}{c}{SADL} \\
\cmidrule(lr){2-3}\cmidrule(lr){4-5}\cmidrule(lr){6-7}\cmidrule(lr){8-9}
Bucket & Dice & LF1 & Dice & LF1 & Dice & LF1 & Dice & LF1 \\
\midrule
$<0.5$~mL         & 0.530 & 0.693 & 0.548 & 0.693 & 0.535 & 0.593 & \textbf{0.592} & \textbf{0.767} \\
$0.5$--$5$~mL     & 0.567 & 0.614 & \textbf{0.596} & 0.666 & 0.575 & 0.642 & 0.579 & \textbf{0.682} \\
$5$--$50$~mL      & 0.764 & 0.684 & 0.759 & 0.727 & 0.768 & 0.699 & \textbf{0.783} & \textbf{0.792} \\
$\geq 50$~mL      & 0.873 & 0.731 & 0.874 & 0.675 & 0.875 & 0.737 & \textbf{0.877} & \textbf{0.756} \\
\bottomrule
\end{tabular}
\end{table}

\setlength{\intextsep}{4pt}
\setlength{\abovecaptionskip}{2pt}
\setlength{\intextsep}{12pt plus 2pt minus 2pt}
\setlength{\abovecaptionskip}{10pt}

A paired non-parametric bootstrap (10{,}000 resamples, BCa-adjusted CIs)
against a reference single member (BucketWeighted with the default
$v_{\text{ref}}=5$~mL) at threshold $0.5$ confirms this pattern; the
bootstrap uses per-case Dice and lesion-F1 because panoptica's global
metrics are aggregated over the whole holdout and are not paired at the
case level. SADL gains $+2.90$~pp in
lesion-F1 with a 95\% CI of $[+0.17, +6.25]$~pp ($p{=}0.024$, significant
at $\alpha{=}0.05$); the Dice gain of $+0.31$~pp is directionally
consistent, but its 95\% CI $[-1.55, +1.72]$~pp includes zero. The
decision layer therefore reliably improves lesion \emph{recognition};
its voxel-Dice effect is holdout-dependent.

\section{Discussion and Conclusion}
\label{sec:discussion}

A well-tuned decision layer is a low-cost gain on top of uniform mean
fusion of a mature nnU-Net ensemble. SADL adds no new training run and
no new member: it re-uses the soft-max outputs already on disk, and
because its parameters are fitted on OOF predictions the tuner is cheap
to re-run whenever a new member joins the pool. Two caveats: our
91-case site-disjoint holdout is not the hidden challenge cohort, and
not every pool strategy paid off (diffusion-augmented synthesis ranked
last despite promising training-time Dice). The three ideas to carry
over are the size-bucket-conditioned threshold, the never-empty rescue
on the handful of $<0.5$~mL cases per fold, and the OOF-fitted fusion
weights. The tuned policy is a small JSON blob that any team running an
nnU-Net ensemble can drop on top of their own fusion, and it drives our
submission container as-is.

\begin{credits}
\vspace{-0.5em}
\subsubsection{\ackname} This work was supported by the National Science
Centre, Poland, under Grant ``MultiGeoMed'' No. 2024/55/D/ST6/02081. We
gratefully acknowledge the Polish high-performance computing infrastructure
PLGrid, HPC Center: ACK Cyfronet AGH, for providing computational resources
and support within computational grant No. PLG/2025/018770. The work was
partially supported by the Excellence Initiative -- Research University
program at the AGH University of Krakow.
\vspace{-0.3em}
\subsubsection{\discintname} The authors have no competing interests to
declare that are relevant to the content of this article.
\end{credits}

{\footnotesize
\vspace{-1em}
\let\OLDthebibliography\thebibliography
\renewcommand{\thebibliography}[1]{%
  \OLDthebibliography{#1}%
  \setlength{\itemsep}{0pt plus 0pt}%
  \setlength{\parsep}{0pt plus 0pt}%
}
\bibliographystyle{splncs04}
\bibliography{references}
}

\end{document}